# Application of machine learning models to predict the relationship between air pollution, ecosystem degradation, and health disparities and lung cancer in Vietnam

Ngoc Hong Tran[1,*], Lan Kim Vien[1], Ngoc-Thao Thi Le[2]

[1]Computer Science University, Vietnamese-German University, Binh Duong, Vietnam

[2]Department of Information Technology, Ho Chi Minh City University of Technology, Ho Chi Minh City, Vietnam

*Corresponding author, Email: ngoc.th@vgu.edu.vn

**ABSTRACT**

Lung cancer is one of the major causes of death worldwide, and Vietnam is not an exception. This disease is the second most common type of cancer globally and the second most common cause of death in Vietnam, just after liver cancer, with 23,797 fatal cases and 26,262 new cases, or 14.4% of the disease in 2020. Recently, with rising disease rates in Vietnam causing a huge public health burden, lung cancer continues to hold the top position in attention and care. Especially together with climate change, under a variety of types of pollution, deforestation, and modern lifestyles, lung cancer risks are on red alert, particularly in Vietnam. To understand more about the severe disease sources in Vietnam from a diversity of key factors, including environmental features and the current health state, with a particular emphasis on Vietnam's distinct socioeconomic and ecological context, we utilize large datasets such as patient health records and environmental indicators containing necessary information, such as deforestation rate, green cover rate, air pollution, and lung cancer risks, that is collected from well-known governmental sharing websites. Then, we process and connect them and apply analytical methods (heatmap, information gain, p-value, spearman correlation) to determine causal correlations influencing lung cancer risks. Moreover, we deploy machine learning (ML) models (Decision Tree, Random Forest, Support Vector Machine, K-mean clustering) to discover cancer risk patterns. Our experimental results, leveraged by the aforementioned ML models to identify the disease patterns, are promising, particularly, the models as Random Forest, SVM, and PCA are working well on the datasets and give high accuracy (99%), however, the K means clustering has very low accuracy (10%) and does not fit the datasets.

## 1. INTRODUCTION

Lung cancer is a leading global health challenge, ranking as the second most common cancer and the foremost cause of cancer-related mortality worldwide. Vietnam, in particular, faces a growing burden of lung cancer, with significant morbidity and mortality rates exacerbated by socioeconomic and environmental factors. In 2020, Vietnam reported over 26,262 new cases and nearly 23,797 deaths due to lung cancer, highlighting the urgent need for effective prevention and intervention strategies (Thai, 2023). According to 2022 data from the Global Cancer Observatory, lung cancer accounted for 13.5% of all cancer cases in Vietnam, with an age-standardized mortality rate of 132.6 per 100,000 people (World Health Organization, 2022).

In the meanwhile, the World Health Organization (WHO) reports that air pollution accounts for 29% of global lung cancer cases, underscoring its role as a major health hazard (WHO, 2024). Deforestation, while indirectly linked, exacerbates respiratory diseases through the loss of carbon sinks and increased particulate matter in the atmosphere (WHO, 2019). Moreover, rapid urbanization and industrialization have intensified environmental degradation in Vietnam (Benschop et al., 2024). The country's reliance on coal for electricity, accounting for 64.6% of generation as of 2024, has significantly increased $CO_2$ emissions. Environmental factors such as air pollution, deforestation, and industrial emissions have been identified as significant contributors to lung cancer. While cigarette smoking remains the primary contributor to lung cancer, accounting for approximately 80% of global cases (Thai, 2023). Beyond smoking, environmental variables such as poor air quality, linked to rising coal consumption and increasing coal use and emissions, are emerging as critical contributors to lung cancer (Ta et al., 2020).

Despite advancements in diagnostic tools and treatments, there remains a lack of comprehensive studies that integrate behavioral and environmental factors, particularly in contexts like Vietnam, where rapid urbanization and industrial growth intersect with public health challenges.

**State of the Art.** Computer-aided diagnosis systems utilizing medical imaging data have been shown to assist physicians in the interpretation of scans, providing a valuable second opinion and improving diagnostic efficiency (Li et al., 2022). Deep learning models, such as Convolutional Neural Networks (CNN), demonstrated promising results in the classification of histopathological lung cancer images, potentially aiding pathologists in the vital process of cancer diagnosis (Baranwal et al, 2021). Ensemble methods, particularly Random Forests, have shown superior accuracy in predicting lung cancer based on genetic, environmental, and lifestyle factors (Oentoro et al, 2023). However, multi-disciplinary works leveraging machine learning to combine different factors including air pollution, climate change via deforestation, and lung risks, are still left for further studies. More studies integrating ML into lung cancer research have primarily focused on predicting cancer risks based on genetic and lifestyle factors. For example, SVMs have demonstrated robust performance in distinguishing between benign and malignant cases based on imaging data (Javed et al., 2024). A 2024 systematic review and meta-analysis highlighted the application of ML models in predicting lung cancer survival rates, emphasizing their potential to enhance diagnostic accuracy and personalized treatment plans (Didier et al., 2024). The review showed that integrating machine learning (ML) techniques in lung cancer analysis has shown promising advancements. However, limited research has applied ML techniques to explore the role of environmental features, such as deforestation and $CO_2$ emissions, in lung cancer incidence (Wang et al., 2022).

In this work, we aim to bridge these gaps by combining different datasets and leveraging data analytics approaches to identify and quantify the significant relationship between environmental and lifestyle factors contributing to lung cancer incidence. We also apply machine learning techniques to predict the lung cancer risk given conditions. Through an analysis of extensive datasets, including patient health records, ecological indicators, and socioeconomic data, this research seeks to uncover actionable insights into the complex interplay of these variables. Focusing on Vietnam's unique context, the findings are intended to inform data-driven policymaking and public health strategies, ultimately aiming to reduce lung cancer's prevalence and societal impact.

The remainder of this paper is organized to 5 sections as follows. The dataset description, integration and preparation are described in Section 2. In Section 3, data analytics and insight findings are presented in details. The classification models and result analytics are demonstrated in Section 4. Eventually, conclusion and future works are presented in Section 5.

## 2. DATASET
### 2.1. Data Sources

In this work, to address the coo relationship among lung cancer, air pollution and forest status, we collected four datasets and combined them to have the final complex data used for analytics. Specially, the four datasets are presented as follows:

**Table 1.** Lung *cancer patient data sets.csv*'s data

| | Patient Id | Age | Gender | Air Pollution | Alcohol use | Dust Allergy | OccuPational Hazards | Genetic Risk | chronic Lung Disease | Balanced Diet | ... | Fatigue | Weight Loss | Shortness of Breath | Wheezing | Swallowing Difficulty | Clubbing of Finger Nails | Frequent Cold | Dry Cough | Snoring | Level |
|---|---|---|---|---|---|---|---|---|---|---|---|---|---|---|---|---|---|---|---|---|---|
| 0 | P1 | 33 | 1 | 2 | 4 | 5 | 4 | 3 | 2 | 2 | ... | 3 | 4 | 2 | 2 | 3 | 1 | 2 | 3 | 4 | Low |
| 1 | P10 | 17 | 1 | 3 | 1 | 5 | 3 | 4 | 2 | 2 | ... | 1 | 3 | 7 | 8 | 6 | 2 | 1 | 7 | 2 | Medium |
| 2 | P100 | 35 | 1 | 4 | 5 | 6 | 5 | 5 | 4 | 6 | ... | 8 | 7 | 9 | 2 | 1 | 4 | 6 | 7 | 2 | High |
| 3 | P1000 | 37 | 1 | 7 | 7 | 7 | 7 | 6 | 7 | 7 | ... | 4 | 2 | 3 | 1 | 4 | 5 | 6 | 7 | 5 | High |
| 4 | P101 | 46 | 1 | 6 | 8 | 7 | 7 | 7 | 6 | 7 | ... | 3 | 2 | 4 | 1 | 4 | 2 | 4 | 2 | 3 | High |

●*Lung Cancer Patient Dataset.* This dataset is sourced from Data.World - Cancer Data HP, provides detailed patient-level information, including demographics, lifestyle factors, medical symptoms, and cancer severity levels (see **Table 1**). Key variables such as smoking status, exposure to passive smoking, air pollution levels, and chronic symptoms like coughing and chest pain enable a granular analysis of risk factors.

**Table 2.** The *Year.csv*'s data

| | Year | Number | Total | Rate |
|---|---|---|---|---|
| 0 | 2000 | 8096 | 68810 | 0.117657 |
| 1 | 2012 | 21865 | 125000 | 0.174920 |
| 2 | 2018 | 23,667 | 164671 | 0.143720 |
| 3 | 2020 | 26262 | 182,563 | 0.143850 |
| 4 | 2022 | 24426 | 180480 | 0.135339 |

**Table 3.** *Forest status (Unit Thous. Ha).csv*'s data

| | Year | Total area of forested land | Natural forest | Planted forest |
|---|---|---|---|---|
| 0 | 2002.0 | 11,784.59 | 9,865.02 | 1,919.57 |
| 1 | 2003.0 | 12,094.52 | 10,004.71 | 2,089.81 |
| 2 | 2004.0 | 12,306.86 | 10,088.29 | 2,218.57 |
| 3 | 2005.0 | 12,616.70 | 10,283.17 | 2,333.53 |
| 4 | 2006.0 | 12,873.85 | 10,410.14 | 2,463.71 |

●*Historical Lung Cancer Cases in Vietnam.* This dataset aggregates longitudinal data on lung cancer incidence in Vietnam from authoritative sources such as Globocan (WHO,

2018) and the Global Cancer Observatory (WHO, 2022). It tracks trends over decades, providing a temporal perspective on lung cancer prevalence and its correlation with evolving environmental conditions (see **Table 2**).

Table 4. The treecover_loss__ha.csv's data

| | iso | umd_tree_cover_loss__year | umd_tree_cover_loss__ha | gfw_gross_emissions_co2e_all_gases__Mg |
|---|---|---|---|---|
| 0 | VNM | 2001 | 47433.13018 | 23021538.75 |
| 1 | VNM | 2002 | 49625.43782 | 23993061.27 |
| 2 | VNM | 2003 | 43007.84845 | 21495654.44 |
| 3 | VNM | 2004 | 73880.88154 | 38621371.50 |
| 4 | VNM | 2005 | 102211.11880 | 54449208.43 |

●*Forest Status Dataset.* Data from Vietnam Forest Change Data (Kiem, 2023) document forest cover trends from 2002 to 2023, capturing changes in natural and planted forests. These records are vital for linking deforestation trends with environmental degradation and its potential impact on lung cancer risk (see **Table 3**).

●*Tree Cover Loss and $CO_2$ Emissions.* Data from Global Forest Watch [14] detail annual tree cover loss and associated $CO_2$ emissions from 2001 to 2023. These metrics directly link deforestation and air quality deterioration, which is essential for assessing environmental contributions to lung cancer incidence (see **Table 4**).

*2.2. Data Integration and Preparation*

To ensure robust analysis, datasets were cleaned, standardized, and integrated for compatibility across diverse sources. Key preprocessing steps included:

●*Categorical Encoding.* Variables such as cancer severity levels were encoded numerically (e.g., "Low," "Medium," and "High" mapped to 1, 2, and 3) to facilitate machine learning model compatibility.

●*Data Cleaning.* To enhance data quality, missing values were imputed, and irrelevant columns were removed. For instance, numerical fields with formatting inconsistencies (e.g., commas) were corrected for computational analysis.

●*Standardization.* All features were scaled to ensure uniformity, using z-scores to standardize values. This step was critical for distance-based algorithms like K-Nearest Neighbors (KNN).

●*Importance of Data Integration.* Integrating micro-level (patient-specific) and macro-level (environmental) data offers a holistic view of lung cancer risk factors. Patient datasets reveal behavioral and genetic influences, while forest and emissions data provide context for ecological degradation. This multi-scale approach enables a nuanced understanding of how environmental and individual factors interact to drive lung cancer prevalence in Vietnam's unique context.

## 3. DATA ANALYTICS AND INSIGHT FINDINGS

In this section, the analytical results are visualized using a heatmap and information gain ranking to identify and analyze the most significant features influencing lung cancer risk.

### *3.1. Heatmap*

Through the heatmap released from the combined dataset (see **Figure 1**), we infer top most impacting features with more than 70% influence ratio connecting the health status and behavior to the lung cancer risks. Particularly, the '**Obesity**' feature affecting 83% is considered as a strong positive correlation, suggesting that higher obesity levels are associated with more severe lung cancer conditions. Secondly, the '**Coughing of Blood**' feature occupying 78% proves that smoking exhibited a significant positive correlation, underscoring its role as a critical factor in lung disease severity. The third-ranked dominant feature is '**Alcohol Use**' with a significance of 72%, indicating a strong correlation with advanced stages of lung cancer. The less dominant features, that are, '**Balanced Diet**' and '**Dust Allergy**' with a 71% influence ratio, highlighting its prevalence in severe cases of lung cancer. '**Balanced Diet**' refers to the situation when the patient lacks some important nutrients while consuming a high amount of food with unnecessary nutrients for their body. Finally, the '**Passive Smoker**' and '**Genetic Risk**' are more impactful than active smoking, passive smoking significantly contributes to disease progression and it reaches a significance of **70%**. Impressively, it becomes evident that even non-smokers have a high probability of developing lung cancer if they are frequently exposed to secondhand cigarette smoke. Ultimately, we must acknowledge that genetics play a significant role in the hereditary transmission of lung cancer. While this realization may be concerning, it also encourages us to prioritize and take better care of our health. Other variables, air pollution, chest pain, and chronic lung diseases demonstrated moderate correlations. These insights emphasize the multifactorial nature of lung cancer risk.

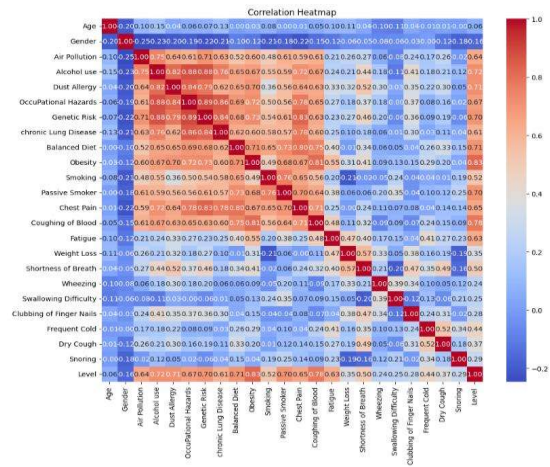

**Figure 1**. Heat map of *cancer patient data sets.csv*

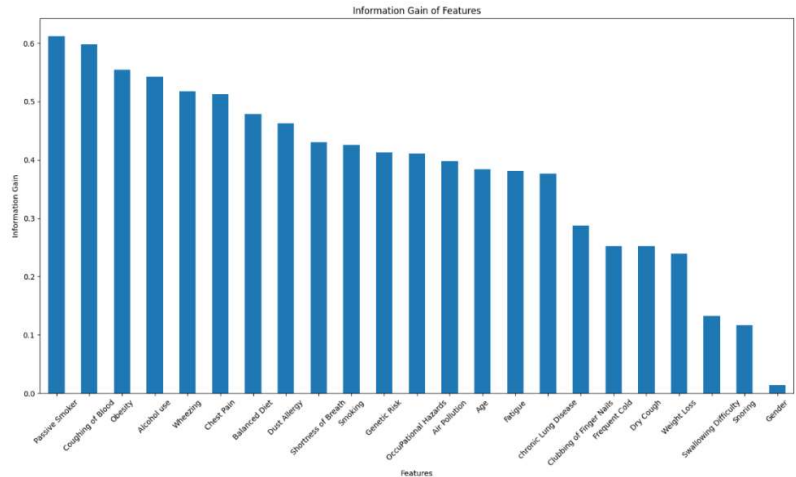

**Figure 2.** Bar chart of information gain of *cancer patient data sets.csv*

## 3.2. Information gain

To support the analytical result from the heatmap above, we use information gain measurement to conclude the dominant features that affect mainly to determine the lung cancer risks at the patients. These findings demonstrate the significance of behavioral and respiratory factors, providing a hierarchy of importance for lung cancer prediction models

The information gain which are achieved in **Figure 2** proves that the top predictors include '**Passive Smoker**' (0.6118), '**Coughing of Blood**' (0.5985), '**Obesity**' (0.5550). The secondly ranked factors are '**Alcohol Use**' (0.5424), '**Wheezing**' (0.5174), '**Chest Pain**' (0.5133). Then, '**Chest Pain**' (0.5133) and '**Balanced Diet**' (0.4782) significantly connect to the lung cancer risks. These ratios and rankings have a high similarity to the results we found in the above heatmap. Additionally, we also find that the least predictive features are '**Gender**' (0.0137), '**Snoring**' (0.1173), and '**Swallowing Difficulty**' (0.1324) contributed minimally to classification performance. Therefore, we can see that if a patient has these symptoms, it does not mean the patients have a lung cancer but the patients have a small probability (approximately 10%) of having lung cancer.

## 3.3. P-value and T-value

**P-Values** is a measure used in hypothesis testing to determine the significance of the results. It represents the probability of obtaining test results at least as extreme as the observed results, assuming that the null hypothesis is true. A lower P-value (typically less than 0.05) suggests that the feature is statistically significant and unlikely to be zero. In other words, the feature has a meaningful relationship with the target. Whereas, **T-Values** is a statistic used in t-tests to determine whether there is a significant difference between the means of two groups. It is calculated as the ratio of the difference between the sample mean and the population mean to the standard error of the sample mean. The T-value helps in determining the P-value, which in turn helps in deciding whether to reject the null hypothesis. Higher absolute T-values (positive or negative) indicate that a feature is statistically significant in predicting the target variable.

Based on T-values and P-values, features like '**Fatigue**', '**Air Pollution**', '**Coughing of Blood**', '**Passive Smoker**', and '**Dry Cough**' are highly significant. They have both high absolute T-values and low P-values, suggesting that they are strongly associated with the target variable. In the meanwhile, the features like '**Snoring**', '**Frequent Cold**', and '**Gender**' have high P-values and low T-values, indicating that they are not statistically significant in predicting the target. These could potentially be dropped from the model to simplify it without much loss in predictive power. For n**egative relationships,** some features, such as '**Dust Allergy**' and '**Age**', have negative T-values, suggesting an inverse relationship with the target. This means that as the values of these features increase, the likelihood or severity of the target condition may decrease (see **Figure 3**).

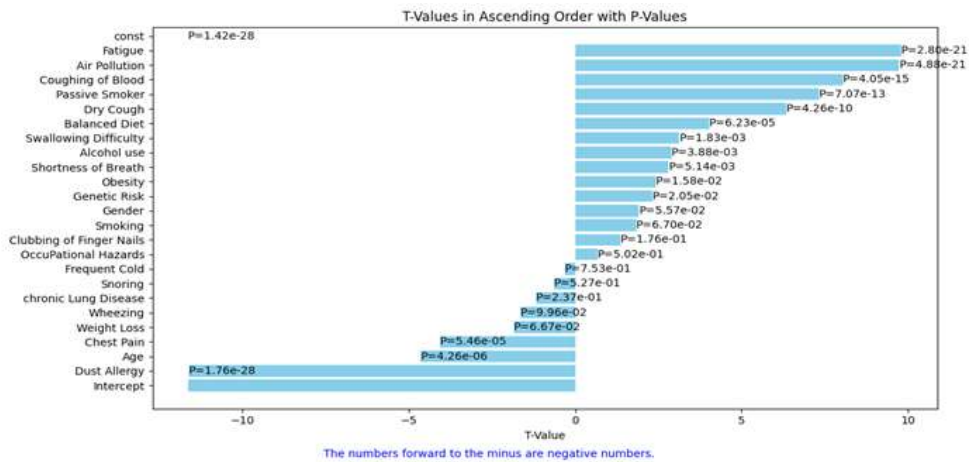

Figure 3. *p*-value and *t*-value in the *cancer patient data sets.csv*

*3.4. Spearman Correlation*

In this section, we aim to measure the correlation between lung cancer risk, deforestation, and air pollution (as indicated by CO2 emission levels) by Spearman correlation. Specifically, we use Spearman correlation to assess their influence on one another. The annual data on tree cover loss is presented in **Figure 5**, which shows an increasing trend. Additionally, **Figure 4** illustrates that both the area of tree cover loss and the number of lung cancer patients are on the rise. The relationship between $CO_2$ emissions from all gases (in Megagrams) and the number of lung cancer patients are observed (see **Figure 6**) and may reflect broader environmental and societal factors rather than a direct cause-and-effect relationship. Therefore, higher $CO_2$ emissions may be associated with increased industrial and transportation activities, which often release other harmful pollutants. These pollutants, rather than $CO_2$ itself, may contribute to lung cancer risk by degrading air quality. Areas with higher $CO_2$ emissions are likely to experience industrial growth and urbanization, which may also lead to an increase in lung cancer diagnoses due to better healthcare infrastructure and more comprehensive reporting of cancer cases.

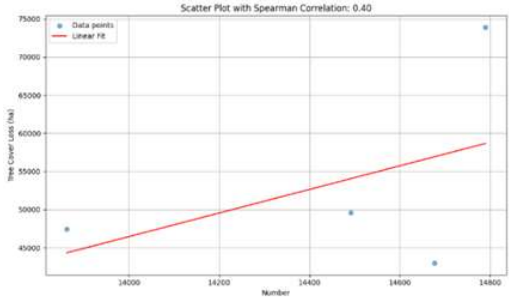 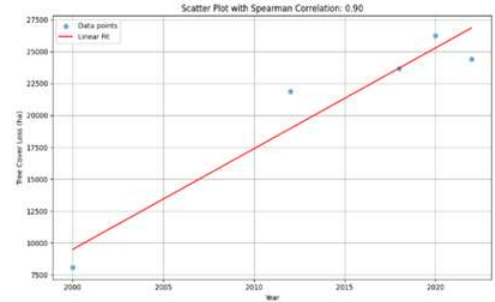

**Figure 4.** Tree Cover Loss (in hectares) and Number of Lung Cancer Patients

**Figure 5.** Year and Tree Cover Loss (in hectares)

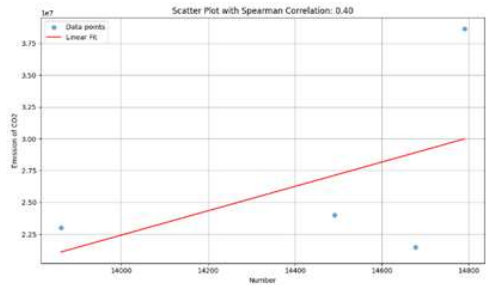

**Figure 6.** $CO_2$ Emissions from All Gases and Number of Lung Cancer Patients

4. **CLASSIFICATION MODELS AND RESULT ANALYTICS**

In this paper, we utilized four machine learning models which are Decision Tree (DT), Random Forest (RF), Support Vector Machine (SVM), and K-means clustering to train and evaluate the data for risk factor prediction.

The dataset is divided into two parts: train and test, with training accounting for 70% and testing for 30%, which included 219 low, 235 medium, and 246 high-risk level data points in training and 84 low, 97 medium, and 119 high-risk level datapoint in test portions. The model training pace is quick; training all four models took about 1-2 seconds.

The classification models were evaluated using precision, recall, and F1 scores to assess their performance. These metrics validate the effectiveness of machine learning models in identifying and classifying lung cancer risk levels with high precision.

**Decision Tree (DT) and Random Forest (RF)**

The DT depicts how the dataset is divided depending on its attributes, with 'Passive Smoker', 'Wheezing', 'Obesity', and 'Snoring' serving as the primary splitting criteria (see **Figure 7**). At each node, a decision is taken based on feature thresholds (e.g., 'Passive Smoker $\leq 6.5$'), and the tree classifies samples into certain classes (Low, Medium, High) as indicated by the leaf nodes.

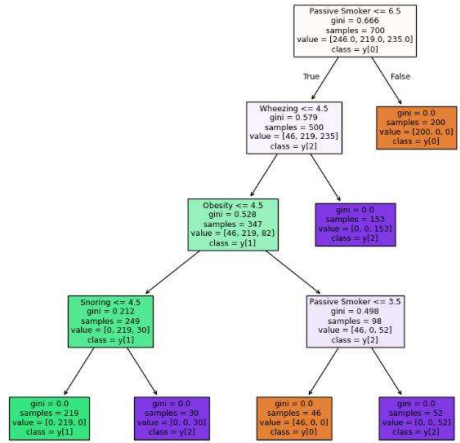

**Figure 7.** Visualization of Decision Tree using the spilt-train method

● *Root Node: The dataset starts with 700 samples and is first split by 'Passive Smoker ≤ 6.5', splitting samples into two groups.*

● *Intermediate Nodes*: Features such as 'Wheezing' and 'Obesity' help to refine the classification, while the Gini index indicates the purity of splits at each node.

● *Leaf Nodes*: Each leaf indicates a pure group, with all samples belonging to the same class (e.g., Low, Medium, or High), as indicated by a Gini index 0.

RF (Subasi, A., 2020) is a collection of DTs. In this experiment, for DT and RF, the prediction is 100% as the number of actual samples matches the predicted samples for all classes: 119 for Low, 84 for Medium, and 97 for High.

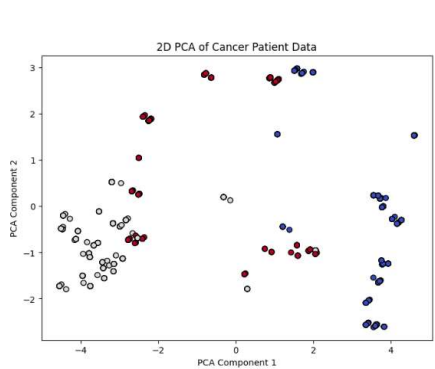
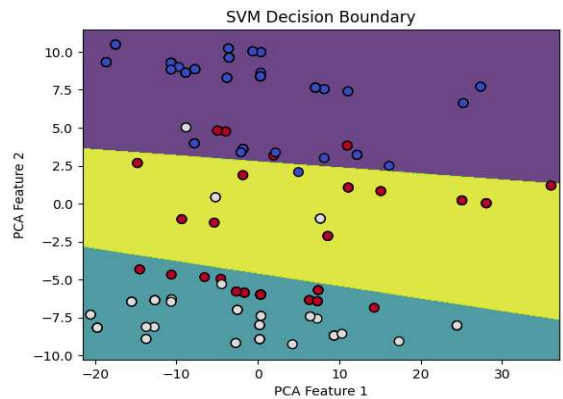

**Figure 8.** SVM's data in the 2D PCA-reduced data

**Figure 9.** SVM's decision boundary

**Support Vector Machine (SVM) and Principle Component Analysis (PCA)**

The dataset likely has many features, making it difficult to visualize in a higher-dimensional space. PCA helps reduce the data into two main components (PCA Component 1 and PCA Component 2) while preserving as much variance in the data as possible. Preparing for SVM to be trained and to learn, we apply PCA to reduce the number of features into 2 PCA. Each point is a patient data plotted in **Figure 8**. Then, we applied the resulted dataset into SVM and the result is demonstrated in **Figure 9**. The different colors (black, red, and blue) likely represent different classes or categories in the dataset, such as different types or stages of cancer or groups related to diagnosis results (e.g., benign, malignant). The experimental result proves that SVM gives a high accuracy as DT and RF, it reaches 100% with these datasets.

**K-means Clustering**

Each dot in **Figure 10** represents a patient in the dataset, and its position is determined by the values of the principal components (after dimensionality reduction with PCA). There are 3 clusters (purple, green, yellow) we use in this method. Each cluster represents patients with similar characteristics (e.g., patients with low cancer risk, patients with medium risk, etc.). These clusters can help understand underlying patterns in the dataset. For example, one cluster could correspond to patients with high air pollution and genetic risk factors, while another could represent patients with lower risk factors.

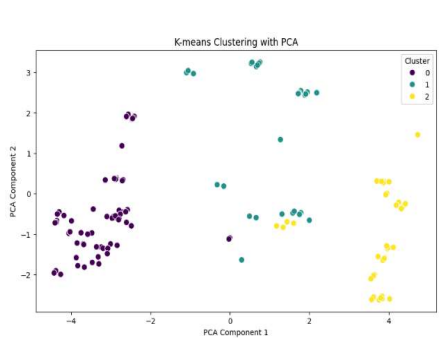

**Figure 10.** K-means clustering with PCA

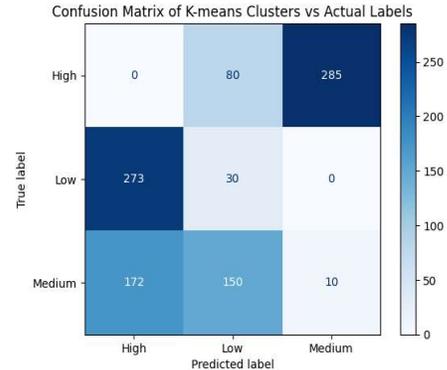

**Figure 11.** Confusion matrix of K-means clustering

We measure the confidence score of K-means clustering method using confusion matrix, however, as visualized in **Figure 11**, the overall clustering accuracy is very low. The number of actual samples poorly matches the predicted samples: none for High (0%), 30 out of 303 samples for Low (10%), and 10 out of 332 samples for Medium (3%). The overall clustering accuracy is very low, indicating significant misclassification by the K-means clustering algorithm.

5. **CONCLUSION**

This study emphasizes the critical role of understanding the interaction between environmental and health factors in the development of lung cancer. Key risk factors, such as air pollution, deforestation and health have been identified. The insight findings can enable public health initiatives to more effectively prevent lung cancer and foster healthier communities. Moreover, the machine learning model is a promising tool to predict, however, we need further deep researches. For future work, we will collect more datasets and leverage more advanced machine models to find out more helpful insights and improve the accuracy of models.